\documentclass{llncs} 

\usepackage{subfig}

\usepackage{tikz}

\usepackage{todonotes}

\begin{document}

\title{\bf Non-monotonic Reasoning\\ in Deductive Argumentation}


\author{Anthony Hunter}

\institute{Department of Computer Science,\\ University College London,\\ London, UK}


\maketitle

\begin{abstract}
Argumentation is a non-monotonic process.
This reflects the fact that argumentation involves uncertain information, and so new information can cause a change in the conclusions drawn.
However, the base logic does not need to be non-monotonic.
Indeed, most proposals for structured argumentation use a monotonic base logic (e.g. some form of modus ponens with a rule-based language, or classical logic).
Nonetheless, there are issues in capturing defeasible reasoning in argumentation
including choice of base logic and modelling of defeasible knowledge.
And there are insights and tools to be harnessed for research in non-monontonic logics.
We consider some of these issues in this paper.
\end{abstract}

\section{Introduction}

Computational  argumentation  is  emerging  as  an  important  part  of
AI research.  This comes from the recognition that if we are
to develop robust intelligent systems, then it is imperative that they can handle incomplete
and inconsistent information in a way that somehow emulates the human ability to tackle such information.  
And one of the key ways that humans do this is to use argumentation, 
either internally, by evaluating arguments and counterarguments, or externally, by for instance
entering into a discussion or debate where arguments are exchanged.
Much research on computational argumentation focuses on one or more of the following layers: 
the structural layer (How are arguments constructed?);
the relational layer (What are the relationships between arguments?); 
the dialogical layer (How can argumentation be undertaken in dialogues?); 
the assessment layer (How can a constellation of interacting arguments be evaluated and conclusions drawn?); 
and  the rhetorical layer (How can argumentation be tailored for an audience so that it is convincing?). 
This has led to the development of a number of formalisms for aspects of argumentation (for reviews see \cite{BH08,RS09,HOFA2018}), and some promising application areas \cite{AIMag2017}).

Argumentation is related to non-monotonic reasoning. 
The latter is reasoning that allows for retraction of inferences in the light of new information. 
Interest in non-monotonic reasoning started with attempts to handle general
rules, or defaults, of the form ``if $\alpha$ holds, then $\beta$
normally holds", where $\alpha$ and $\beta$ are propositions.  It is
noteworthy that human practical reasoning relies much more on
exploiting general rules (not to be understood as universal laws) than
on a myriad of individual facts. General rules tend to be less than
100\% accurate, and so have exceptions. Nevertheless it is intuitive
and advantageous to resort to such defaults and therefore allow the
inference of useful conclusions, even if it does entail making some
mistakes as not all exceptions to these defaults are necessarily
known. Furthermore, it is often necessary to use general rules when we
do not have sufficient information to allow us to specify or use
universal laws. For a review of non-monotonic reasoning, see \cite{Brewka1991}.

In using defeasible (or default) knowledge, we might make an inference $\alpha$ on the basis of the information available, and then on the basis of further information, we may want to withdraw $\alpha$. So with defeasible knowledge, the set of inferences does not increase monotonically with the set of assumptions.

\begin{example}
Consider the following general statements
with the fact {\tt the match is struck}. For this, we infer,
$\mbox{\tt The match lights}$. 
But, if we also have the fact
$\mbox{\tt The match is wet}$
then we retract
$\mbox{\tt The match lights}$. 
\[
\mbox{\tt A match lights if struck}
\]
\[
\mbox{\tt A match doesn't light if struck}
\]
\end{example}

The notion of defeasible knowledge covers a diverse variety of
information, including heuristics, rules of conjecture, null values in
databases, closed world assumptions for databases, and some
qualitative abstractions of probabilistic information. Defeasible knowledge is a
natural and very common form of information. There are also obvious
advantages to applying the same default a number of times: There is an
economy in stating (and dealing with) only a general rule instead of
stating (and dealing with) many refined instances of such a
general rule.

Even though argumentation and non-monotonic reasoning are widely acknowledged as closely related phenomena, there is a need to clarify the relationship. 
There is pioneering work such as by Simari {\em et. al.} \cite{SL92,GS04}, Prakken {\em et. al.} \cite{Prakken1993,PS97,Prakken10}, Bondarenko \cite{Bon97}, and Toni \cite{Toni2013,Toni14} that looks at aspect of this relationship, but it is potentially valuable to revisit the topic, and in particular investigate it from the point of view of deductive argumentation.  
So our primary aim in this paper is to investigate how non-monotonic reasoning arises in deductive argumentation, and how that differs from logics for non-monotonic reasoning. Our secondary aim is to investigate how logics for non-monotonic reasoning can be harnessed in deductive argumentation.

We proceed in the rest of the paper as follows:
In Section \ref{section:deductive}, 
we review deductive argumentation as an example of a framework for structured argumentation; 
In Section \ref{section:nmr},
we consider how non-monotonic reasoning arises in structured argumentation; 
In Section \ref{section:default},
we review default logic, and consider how it can be harnessed in deductive argumentation; 
In Section \ref{section:conditional},
we review conditional logics, and consider how they can be harnessed in deductive argumentation; 
In Section \ref{section:modelling},
we consider how we can model defeasible knowledge as defeasible rules; 
And in Section \ref{section:conclusions}, we draw conclusions and consider future work.

\section{Deductive argumentation}
\label{section:deductive}

Deductive argumentation is formalized in terms of deductive arguments and counterarguments,
and there are various choices for defining this \cite{BH08,BH14}.
In the rest of this section, we will investigate some of the choices we have for defining arguments and counterarguments, and for how they can be used in modelling argumentation. 

In order to define a specific system for deductive argumentation, we need to use a base logic. This is a logic that specifies the logical language for the knowledge, and the consequence (or entailment) relation for deriving inferences from the knowledge. 
In this section, we focus on two choices for base logic. These are simple logic (which has a language of literals and rules of the form $\alpha_1\wedge\ldots\wedge\alpha_n\rightarrow\beta$ where $\alpha_1,\ldots,\alpha_n,\beta$ are literals, and modus ponens is the only proof rule) and classical logic (propositional and first-order classical logic).

A {\bf deductive argument} is an ordered pair $\langle\Phi,\alpha\rangle$ where $\Phi\vdash_i\alpha$ holds for the base logic $\vdash_i$.
$\Phi$ is the support, or premises, or assumptions of the argument, and $\alpha$ is the claim, or conclusion, of the argument.
Different deductive argumentation systems can be obtained by imposing constraints (such as minimality or consistency) on the definition of an argument. 
For an argument $A = \langle \Phi, \alpha \rangle$, 
the function ${\sf Support}(A)$ returns $\Phi$ 
and the function ${\sf Claim}(A)$ returns $\alpha$. 

A counterargument is an argument that attacks another argument. In deductive argumentation, we define the notion of counterargument in terms of logical contradiction between the claim of the counterargument and the premises or claim of the attacked argument. We will review some of the kinds of counterargument that can be specified for simple logic and classical logic.

\subsection{Simple logic}

Simple logic is based on a language of literals and simple rules where each {\bf simple rule} is of the form $\alpha_1 \wedge \ldots \wedge \alpha_k \rightarrow \beta$ where $\alpha_1$ to $\alpha_k$ and $\beta$ are literals. A {\bf simple logic knowledgebase} is a set of literals and a set of simple rules. The consequence relation is modus ponens (i.e. implication elimination).

\begin{definition}
The {\bf simple consequence relation}, denoted $\vdash_s$, 
which is the smallest relation satisfying the following condition, 
and where $\Delta$ is a simple logic knowledgebase: 
$\Delta\vdash_s\beta$ iff 
there is an $\alpha_1 \wedge \cdots \wedge \alpha_n \rightarrow \beta$ $\in$ $\Delta$,
and for each $\alpha_i \in \{\alpha_1,\ldots,\alpha_n\}$, 
either $\alpha_i  \in \Delta$ or $\Delta\vdash_s\alpha_i$.
\end{definition}

\begin{example}
Let $\Delta = \{ a, b, a \wedge b \rightarrow c, c \rightarrow d \}$.
Hence, $\Delta \vdash_s c$ and $\Delta \vdash_s d$.
However, $\Delta \not\vdash_s a$ and $\Delta \not\vdash_s b$.
\end{example}

\begin{definition}
Let $\Delta$ be a simple logic knowledgebase. 
For $\Phi \subseteq\Delta$,
and a literal $\alpha$,
$\langle \Phi, \alpha \rangle$ is a {\bf simple argument}
iff $\Phi \vdash_s \alpha$
and there is no proper subset $\Phi'$ of $\Phi$ 
such that $\Phi' \vdash_s \alpha$. 
\end{definition}

So each simple argument is minimal but not necessarily consistent (where consistency for a simple logic knowledgebase $\Delta$ means that for no atom $\alpha$ does $\Delta\vdash_s\alpha$ and $\Delta\vdash_s\neg\alpha$ hold). We do not impose the consistency constraint in the definition for simple arguments as simple logic is paraconsistent, and therefore can support a credulous view on the arguments that can be generated. 

\begin{example}
Let ${\tt p_1}$, ${\tt p_2}$, and ${\tt p_3}$ be the following formulae.
Note, we use ${\tt p_1}$, ${\tt p_2}$, and ${\tt p_3}$ as labels in order to make the presentation of the premises more concise.		
Then $\langle \{\tt p_1, p_2, p_3 \},{\tt goodEmployee(John)}	\rangle$ is a simple argument.
\[
\begin{array}{l}
{\tt p_1} = {\tt clever(John)}\\
{\tt p_2} = {\tt conscientious(John)}\\
{\tt p_3} = {\tt clever(John)} \wedge {\tt conscientious(John)} \rightarrow {\tt goodEmployee(John)}
\end{array}
\]		
\end{example}

For simple logic, we consider two forms of counterargument. For this, recall that literal $\alpha$ is the complement of literal $\beta$ if and only if $\alpha$ is an atom and $\beta$ is $\neg\alpha$ or if $\beta$ is an atom and $\alpha$ is $\neg\beta$.

\begin{definition}
For simple arguments $A$ and $B$,  we consider the following type of {\bf simple attack}:
\begin{itemize}
\item $A$ is a {\bf simple undercut} of $B$ if there is a simple rule $\alpha_1 \wedge \cdots \wedge \alpha_n \rightarrow \beta$ in ${\sf Support}(B)$
and there is an $\alpha_i \in \{\alpha_1,\ldots,\alpha_n\}$
such that ${\sf Claim}(A)$ is the complement of $\alpha_i$.
\item $A$ is a {\bf simple rebut} of $B$ if ${\sf Claim}(A)$ is the complement of ${\sf Claim}(B)$.
\end{itemize}
\end{definition}

\begin{example}
\label{ex:counterarguments:simple:metro1}
The first argument $A_1$ captures the reasoning that the metro is an efficient form of transport, so one can use it. The second argument $A_2$ captures the reasoning that there is a strike on the metro, and so the metro is not an efficient form of transport (at least on the day of the strike). $A_2$ undercuts $A_1$.
\[
\begin{array}{l}
A_1 = {\tt \langle \{ efficientMetro, efficientMetro \rightarrow useMetro \}, useMetro \rangle}\\
A_2 = {\tt \langle \{ strikeMetro, strikeMetro \rightarrow \neg efficientMetro \}, \neg efficientMetro \rangle}\\
\end{array}
\]
\end{example}

\begin{example}
\label{ex:counterarguments:simple:gov1}
The first argument $A_1$ captures the reasoning that the government has a budget deficit, and so the government should cut spending. The second argument  $A_2$ captures the reasoning that the economy is weak, and so the government should not cut spending. The arguments rebut each other.
\[
\begin{array}{l}
A_1 = {\tt \langle \{ govDeficit, govDeficit \rightarrow cutGovSpending \}, cutGovSpending \rangle} \\
A_2 = {\tt \langle \{ weakEconomy, weakEconomy \rightarrow \neg cutGovSpending \}, \neg cutGovSpending \rangle}\\
\end{array}
\]
\end{example}

So in simple logic, a rebut attacks the claim of an argument, and an undercut attacks the premises of the argument (either by attacking one of the literals, or by attacking the consequent of one of the rules in the premises).

\subsection{Classical logic}

Classical logic is appealing as the choice of base logic as it better reflects the richer deductive reasoning often seen in arguments arising in discussions and debates. 

We assume the usual propositional and predicate (first-order) languages for classical logic, and the usual 
the {\bf classical consequence relation}, denoted $\vdash$.
A {\bf classical knowledgebase} is a set of classical propositional or predicate formulae.

\begin{definition}
For a classical knowledgebase $\Phi$,
and a classical formula $\alpha$,
$\langle \Phi, \alpha \rangle$ is a {\bf classical argument}
iff $\Phi \vdash \alpha$ 
and $\Phi \not\vdash\bot$
and there is no proper subset $\Phi'$ of $\Phi$ 
such that $\Phi' \vdash \alpha$. 
\end{definition}

So a classical argument satisfies both minimality and consistency. We impose the consistency constraint because we want to avoid the useless inferences that come with inconsistency in classical logic (such as via ex falso quodlibet).

\begin{example}
The following classical argument uses a universally quantified formula in contrapositive reasoning to obtain the  claim about number 77.
\[
\tt \langle \{ \forall X. multipleOfTen(X) \rightarrow even(X), \neg even(77) \}, \neg multipleOfTen(77)  \rangle
\]
\end{example}

Given the expressivity of classical logic (in terms of language and inferences), there are a number of natural ways that we can define counterarguments. We give some options in the following definition. 

\begin{definition}
\label{def:classical:attack}
Let $A$ and $B$ be two classical arguments. 
We define the following types of {\bf classical attack}.
\[
\begin{array}{l}
A \mbox{ is a {\bf classical undercut} of} B \mbox{ if } 
\exists \Psi \subseteq {\sf Support}(B) 
\mbox{ s.t. } {\sf Claim}(A) \equiv \neg \bigwedge \Psi.\\
A \mbox{ is a {\bf classical direct undercut} of } B 
\mbox{ if }  \exists \phi \in {\sf Support}(B) \mbox{ s.t. }  {\sf Claim}(A) \equiv \neg \phi.\\
A \mbox{ is a {\bf classical rebuttal} of } B \mbox{ if }  {\sf Claim}(A) \equiv \neg {\sf Claim}(B).\\
\end{array}
\]
\end{definition}

Using simple logic, the definitions for counterarguments against the support of another argument are limited to attacking just one of the items in the support. In contrast,  using classical logic, a counterargument can be against more than one item in the support. For example, in Example \ref{ex:classical:counterarguments:airline}, the undercut is not attacking an individual premise but rather saying that two of the premises are incompatible (in this case that the premises ${\tt lowCost}$ and ${\tt luxury}$ are incompatible).

\begin{example}
\label{ex:classical:counterarguments:airline}
Consider the following arguments.
$A_1$ is attacked by $A_2$ as $A_2$ is an undercut of $A_1$ though it is not a direct undercut.
Essentially, the attack is an integrity constraint.
\[
\begin{array}{l}
A_1 = \langle \{ {\tt lowCost}, {\tt luxury}, {\tt lowCost} \wedge {\tt luxury} \rightarrow {\tt goodFlight} \}, {\tt goodFlight} \rangle\\						
A_2 = \langle \{ \neg {\tt lowCost} \vee \neg {\tt luxury} \}, \neg {\tt lowCost} \vee \neg {\tt luxury} \rangle\\
\end{array}
\]
\end{example}

We give further examples of undercuts in Figure \ref{fig:classical:medical}.

\subsection{Instantiating argument graphs}

For a specific deductive argumentation system, 
once we have a definition for arguments, and for counterarguments (i.e. for the attack relation),  we can consider how to use them to instantiate argument graphs. For this, we need to specify which arguments and attacks are to appear in the instantiated argument graph.  Two approaches to specifying this are descriptive graphs and generative graphs defined informally as follows.

\begin{itemize}

\item {\bf Descriptive graphs} Here we assume that the structure of the argument graph is given, and the task is to identify the premises and claim of each argument. Therefore the input is an abstract argument graph, and the output is an instantiated argument graph. This kind of task arises in many situations: For example,  if we are listening to a debate, we hear the arguments exchanged, and we can construct the instantiated argument graph to reflect the debate.

\item {\bf Generative graphs} Here we assume that we start with a knowledgebase (i.e. a set of logical formula), and the task is to generate the arguments and counterarguments (and hence the attacks between arguments). Therefore, the input is a knowledgebase, and the output is an instantiated argument graph. 
This kind of task also arises in many situations: For example, if we are making a decision based on conflicting information. We have various items of information that we represent by formulae in the knowledgebase, and we construct an instantiated argument graph to reflect the arguments and counterarguments that follow from that information. 
Note, we do not need to include all the arguments and attacks that we can generate from the knowledgebase. Rather we can define a selection function to choose which arguments and attacks to include. 

\end{itemize}

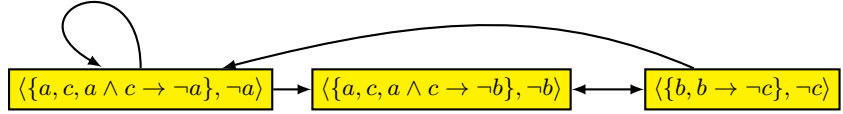
\begin{figure}[t]
\begin{center}
\begin{tikzpicture}[>=latex,thick,every node/.style={draw,fill=yellow,rectangle},scale=0.8]
\node (A1)  [] at (0,0.5) {$\langle \{ a,c, a \wedge c \rightarrow \neg a \}, \neg a \rangle$};
\node (A2)  [] at (5,0.5) {$\langle \{ a,c, a \wedge c \rightarrow \neg b \}, \neg b \rangle$};
\node (A3)  [] at (10,0.5) {$\langle \{ b, b \rightarrow \neg c \}, \neg c \rangle$};
\path (A1) edge[->] (A2);
\path (A2) edge[<->] (A3);
\draw[->] (A3) .. controls (7,1.8) and (5,1.8) .. (A1);
\path (A1) edge[->,in=150,out=90,loop] (AA1);
\end{tikzpicture}
\end{center}
\caption{\label{fig:simple:selfcycle}
A generative graph obtained from the simple logic knowlegebase 
where $\Delta = \{ a, b, c, 
	a \wedge c \rightarrow \neg a, 
	b \rightarrow \neg c, 
	a \wedge c \rightarrow \neg b \}$.
Note, that this exhaustive graph contains a self cycle, and an odd length cycle. }
\end{figure}

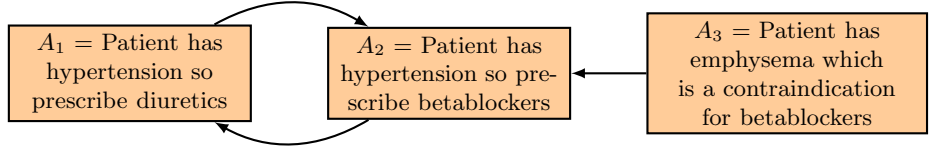
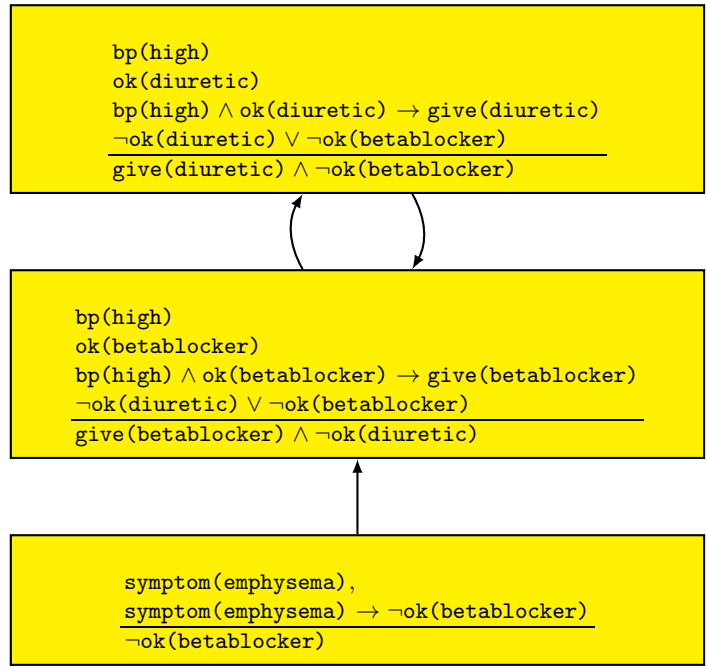
\begin{figure}%
\centering
\subfloat[Argument graph.]
{
\begin{tikzpicture}[->,>=latex,thick]
\node (a1) [text centered,text width=3cm,shape=rectangle,fill=orange!40,draw] {$A_1$ = Patient has hypertension so prescribe diuretics};
\node (a2) [right=of a1,text centered,text width=3cm,shape=rectangle,fill=orange!40,draw] {$A_2$ = Patient has hypertension so prescribe betablockers};
\node (a3) [right=of a2,text centered,text width=3.5cm,shape=rectangle,fill=orange!40,draw] {$A_3$ = Patient has emphysema which is a contraindication for betablockers};
\path	(a1)[bend left] edge node[auto] {} (a2);
\path	(a2)[bend left] edge node[] {} (a1);
\path	(a3) edge node[] {} (a2);
\end{tikzpicture}
}\\
\subfloat[Descriptive graph representation of the argument graph.]
{
\begin{tikzpicture}[->,>=latex,thick]
\node (a1) [text centered,text width=9cm,shape=rectangle,fill=yellow,draw] {
						\[
						\begin{array}{l}
						\mbox{\small \tt bp(high)}\\
						\mbox{\small \tt ok(diuretic)}\\
						\mbox{\small \tt bp(high)} \wedge \mbox{\small \tt ok(diuretic)} \rightarrow \mbox{\small \tt give(diuretic)}\\
						\neg \mbox{\small \tt ok(diuretic)} \vee \neg \mbox{\small \tt ok(betablocker)}\\
						\hline
						\mbox{\small \tt give(diuretic)} \wedge \neg \mbox{\small \tt ok(betablocker)}						
						\end{array}
						\]};
			\node (a2) [below=of a1,text centered,text width=9cm,shape=rectangle,fill=yellow,draw] {
						\[
						\begin{array}{l}
						\mbox{\small \tt bp(high)} \\
						\mbox{\small \tt ok(betablocker)}\\
						\mbox{\small \tt bp(high)} \wedge \mbox{\small \tt ok(betablocker)} \rightarrow \mbox{\small \tt give(betablocker)}\\
						\neg \mbox{\small \tt ok(diuretic)} \vee \neg \mbox{\small \tt ok(betablocker)}\\
						\hline
						\mbox{\small \tt give(betablocker)} \wedge \neg \mbox{\small \tt ok(diuretic)}
						\end{array}
						\]};
			\node (a3) [below=of a2,text centered,text width=9cm,shape=rectangle,fill=yellow,draw] {
						\[
						\begin{array}{l}
						\mbox{\small \tt symptom(emphysema)}, \\
						\mbox{\small \tt symptom(emphysema)} \rightarrow \neg \mbox{\small \tt ok(betablocker)} \\
						\hline
						\neg \mbox{\small \tt ok(betablocker)}
						\end{array}
						\]};
			\path	(a1)[bend left] edge node[auto] {} (a2);
			\path	(a2)[bend left] edge node[] {} (a1);
			\path	(a3) edge node[] {} (a2);
\end{tikzpicture}
}\\
\caption{\label{fig:classical:medical}
The abstract argument graph captures a decision making scenario where there are two alternatives for treating a patient, diuretics or betablockers. Since only one treatment should be given for the disorder, each argument attacks the other. There is also a reason to not give betablockers, as the patient has emphysema which is a contraindication for this treatment.
The descriptive graph representation of the abstract argument graph is using classical logic. The atom ${\tt bp(high)}$ denotes that the patient has high blood pressure.  The top two arguments rebut each other (i.e. the attack is classical defeating rebut). For this, each argument has an integrity constraint in the premises that says that it is not ok to give both betablocker and diuretic. So the top argument is attacked on the premise ${\tt ok(diuretic)}$ and the middle argument is attacked on the premise ${\tt ok(betablocker)}$.}
\label{3figs}
\end{figure}

We give an example of generative graph in Figure \ref{fig:simple:selfcycle} and an example of a descriptive graph in Figure \ref{fig:classical:medical}.

For constructing both descriptive graphs and generative graphs, there may be a dynamic aspect to the process. For instance, when constructing descriptive graphs, we may be unsure of the exact structure of the argument graph, and it is only by instantiating individual arguments that we are able to say whether it is attacked or attacks another argument. As another example, when constructing generative graphs, we may be involved in a dialogue, and so through the dialogue, we may obtain further information which allows us to generate further arguments that can be added to the argument graph.

\subsection{Deductive argumentation as a framework}

\begin{figure}[t]
\begin{center}
\begin{tikzpicture}[->,>=latex,thick]
\node (n2) [text centered,text width=7.5cm,shape=rectangle,fill=gray!30,draw] at (4.5,3.2) {Evaluation criteria};
\node (5) [text centered,text width=3.5cm,shape=rectangle,fill=magenta!40,draw] at (6.5,2.4) {Generative graphs};
\node (4) [text centered,text width=3.5cm,shape=rectangle,fill=magenta!40,draw] at (2.5,2.4) {Descriptive graphs};
\node (3) [text centered,text width=7.5cm,shape=rectangle,fill=magenta!40,draw] at (4.5,1.6) {Counterarguments};
\node (2) [text centered,text width=7.5cm,shape=rectangle,fill=magenta!40,draw] at (4.5,0.8) {Arguments};
\node (1) [text centered,text width=7.5cm,shape=rectangle,fill=magenta!40,draw] at (4.5,0) {Base logic};
\node (n1) [text centered,text width=7.5cm,shape=rectangle,fill=gray!30,draw] at (4.5,-0.8) {Knowledgebase};
\end{tikzpicture}
\end{center}
\caption{\label{fig:framework}Framework for constructing argument graphs with deductive arguments: 
For defining a specific argumentation system, there are four levels for the specification: 
(1) A base logic is required for defining the logical language and the consequence or entailment relation (i.e. what inferences follow from a set of formlulae); 
(2) A definition of an argument $\langle\Phi,\alpha\rangle$ specified using the base logic (e.g. $\Phi$ is consistent, and $\Phi$ entails $\alpha$); 
(3) A definition of counterargument specified using the base logic (i.e. a definition for when one argument attacks another); 
and (4) A definition of which arguments and counterarguments are composed into an argument graph (which is either a descriptive graph or some form of generative graph).  Then to use a deductive argumentation system, a knowledgebase needs to be specified in the language of the base logic, and evaluation criteria such as dialectical semantics need to be selected.}
\end{figure}
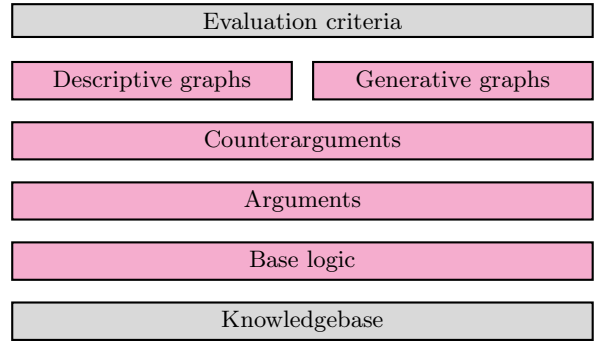

So in order to construct argument graphs with deductive arguments, we need to specify the the base logic, the definition for arguments, the definition for counterarguments, and the definition for instantiating argument graphs. For the latter, we can either produce a descriptive graph or a generative graph.  We summarize the framework for constructing argument graphs with deductive arguments in Figure \ref{fig:framework}.

Key benefits of deductive arguments include: (1) Explicit representation of the information used to support the claim of the argument; (2) Explicit representation of the claim of the argument; and (3) A simple and precise connection between the support and claim of the argument via the consequence relation. What a deductive argument does not provide is a specific proof of the claim from the premises. There may be more than one way of proving the claim from the premises, but the argument does not specify which is used. It is therefore indifferent to the proof used.

There are a number of proposals for deductive arguments using classical propositional logic \cite{Cayrol95,BH01,AC02,GH11}, classical predicate logic \cite{BH05}, description logic \cite{BHP09,MWF10,ZZXL10,ZL13}) temporal logic \cite{MH08}, simple (defeasible) logic \cite{GMAB04,Hun10base}, conditional logic \cite{BGR13}, and probabilistic logic \cite{Hae98,Hae01,Hun13}. 
These are monotonic logics, though non-monotonic logics can be used as a base logics, as we will investigate in this paper.

There has also been progress in understanding the nature of classical logic in computational argumentation.
Types of counterarguments
include rebuttals \cite{Pollock87,Pollock92}, 
direct undercuts \cite{Elvang93,Elvang95,Cayrol95}, 
and undercuts and canonical undercuts \cite{BH01}. 
In most proposals for deductive argumentation, an argument $A$ is a counterargument to an argument $B$ when the claim of $A$ is inconsistent with the support of $B$. It is possible to generalize this with alternative notions of counterargument. For instance, with some common description logics, there is not an explicit negation symbol. In the proposal for argumentation with description logics, \cite{BHP09} used the description logic notion of {\em incoherence} to define the notion of counterargument: A set of formulae in a description logic is incoherent when there is no set of assertions (i.e. ground literals) that would be consistent with the formulae. Using this, an argument $A$ is a counterargument to an argument $B$ when the claim of $A$ together with the support of $B$ is incoherent.

\section{Non-monotonic reasoning}
\label{section:nmr}

For a logic with a consequence relation $\vdash_i$, an important property is the monotonicity property (below) which states that if $\alpha$ follows from a knowledgebase $\Delta$, it still follows from $\Delta$ augmented with any additional formula. This is a property that holds for many logics including classical logic, intuitionistic logic, and many modal and temporal logics.  

\[
\frac{\Delta\vdash_i\alpha}{\Delta\cup\{\beta\}\vdash_i\alpha} \hspace{1cm} \mbox{[Monotonicity]}
\]

In the following example, we assume a consequence relation $\vdash_i$ for which the monotonicity property does not hold, and show how it reflects a form of defeasible reasoning.  

\begin{example}
Consider the following set of formulae $\Delta$ where $\Rightarrow$ is some form of conditional implication symbol. 
\[
\begin{array}{l}
{ \tt bird(x) \Rightarrow \mbox{\tt flyingThing}(x)} \\
{ \tt ostrich(x) \Rightarrow \neg \mbox{\tt flyingThing}(x) }\\
{ \tt ostrich(x) \Rightarrow bird(x)} \\
\end{array}
\]
Assuming we have an appropriate non-monotonic logic, we would get the following behaviour. So given the rules in $\Delta$ and the premise ${\tt bird(Tweety)}$ we get $\mbox{\tt flyingThing(Tweety)}$, but when we add ${\tt ostrich(Tweety)}$ to the premises, we no longer get $\mbox{\tt flyingThing(Tweety)}$. 
\[
\begin{array}{c}
{ \tt \Delta\cup \{ bird(Tweety) \} \vdash_i \mbox{\tt flyingThing(Tweety)}}\\
{ \tt \Delta\cup \{ bird(Tweety),ostrich(Tweety) \} \not\vdash_i \mbox{\tt flyingThing}(Tweety)}\\
\end{array}
\]
In the following sections, we will consider specific logics that give us such behaviour. 
\end{example}

\subsection{Non-monotonicity in deductive argumentation}

We now focus on deductive argumentation, and investigate the monotonic and non-monotonic aspects of it. 

\begin{description}

\item[Argument construction is monotonic] In deductive reasoning, we start with some premises, and we derive a conclusion using one or more inference steps in a base logic.
Within the context of an argument, if we regard the premises as credible, then we should regard the intermediate conclusion of each inference step as credible, and therefore we should regard the conclusion as credible. 
For example, if we regard the premises ``Philippe and Tony are having tea together in London" as credible, then we should regard that ``Philippe is not in Toulouse" as credible (assuming the background knowledge that London and Toulouse are different places, and that nobody can be in different places at the same time). 
As another example, if we regard that the statement ``Philippe and Tony are having an ice cream together in Toulouse" is credible, then we should regard the statement ``Tony is not in London" as credible.
Note, however, we do not need know that the premises are true to apply deductive reasoning. 
Rather, deductive reasoning allows us to obtain conclusions that we can regard as credible contingent on the credibility of their premises.
This means that we can form arguments monotonically from a knowledgebase: Adding formulae to the knowledgebase allows us to increase the set of arguments. 

\item[Argument evaluation is non-monotonic]
Given a knowledgebase, we use the base logic to construct the arguments and to identify the attack relationships that hold between arguments. We then apply the dialectical criteria to determine which arguments are in an extension according to a specific semantics. If we then add more formulae to the knowledgebase, we may obtain further arguments, but we do not lose any arguments. So when we add formulae to the knowledgebase, we may make additions to the instantiated argument graph but we do not make any deletions. If we then apply the dialectical criteria, we may then lose arguments from our extension. So in this sense, argumentation is non-monotonic.

\end{description}

 In the following example, we illustrate how argument construction is monotonic but argument evaluation is non-monotonic.

\begin{example}
\label{ex:argnm}
Let $\Delta = \{ a \}$, and so the following is a generative argument graph where $A'_1$ stands for all arguments with claim implied by $a$ such as $\langle \{ a \},a \vee b \rangle$, $\langle \{ a \},a \vee c \rangle$, etc.
Hence, $\{ A_1, A'_1,\ldots \}$ is the grounded extension of the graph.
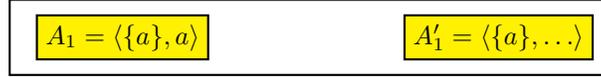
\begin{center}
\begin{tikzpicture}[>=latex,->,thick,every node/.style={draw,rectangle,fill=yellow}]
			\node (1) at (3.5,0) {$A_1 = \langle \{ a \}, a \rangle$};
			\node (4) at (8.5,0) {$A'_1 = \langle \{ a \}, \ldots \rangle$};			
\draw (2,-0.5) rectangle (10,0.5);
\end{tikzpicture}
\end{center}
If we add $\neg a$ to $\Delta$, we get the following generative argument graph
where $A'_2$ stands for all arguments with claim implied by $\neg a$ such as $\langle \{ \neg a \}, \neg a \vee b \rangle$, $\langle \{ \neg a \}, \neg a \vee c \rangle$, etc.
Hence, $A_1$ is no longer in the grounded extension.
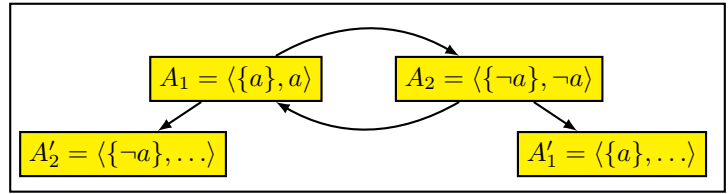
\begin{center}
\begin{tikzpicture}[>=latex,->,thick,every node/.style={draw,rectangle,fill=yellow}]
			\node (1) at (3.5,1) {$A_1 = \langle \{ a \}, a \rangle$};
			\node (2) at (7,1) {$A_2 = \langle \{ \neg a \}, \neg a \rangle$};
			\node (3) at (2,0) {$A'_2 = \langle \{ \neg a \}, \ldots \rangle$};
			\node (4) at (8.5,0) {$A'_1 = \langle \{ a \}, \ldots \rangle$};			
			\path	(1)[bend left] edge (2);
			\path	(2)[bend left] edge (1);
			\path	(1)[] edge (3);
			\path	(2)[] edge (4);		
\draw (0.5,-0.5) rectangle (10,2);

\end{tikzpicture}
\end{center}
\end{example}

Simple arguments and counterarguments can be used to model defeasible reasoning. For this, we use simple rules that are normally correct but sometimes are incorrect. For instance, if Sid has the goal of going to work, Sid takes the metro. This is generally true, but sometimes Sid works at home, 
and so it is no longer true that Sid takes the metro, as we see in the next example.

\begin{example}
\label{ex:counterarguments:simple:metro2}
The first argument $A_1$ captures the general rule that if ${\tt workDay}$ holds, then ${\tt useMetro(Sid)}$ holds. The use of the simple rule in $A_1$ requires that the assumption ${\tt normal}$ holds. This is given as an assumption. The second argument $A_2$ undercuts the first argument by contradicting the assumption that ${\tt normal}$ holds
\[
\begin{array}{l}
A_1 = {\tt \langle \{ workDay, normal, workDay \wedge normal \rightarrow useMetro(Sid) \}, useMetro(Sid) \rangle} \\
A_2 = {\tt \langle \{ workAtHome(Sid), workAtHome(Sid) \rightarrow \neg normal \}, \neg normal \rangle}
\end{array}
\]
If we start with just argument $A_1$, then $A_1$ is undefeated, and so ${\tt useMetro(Sid)}$ is an acceptable claim. However, if we add $A_2$, then $A_1$ is a defeated argument and $A_2$ is an undefeated argument. Hence, if we have $A_2$, we have to withdraw ${\tt useMetro(Sid)}$ as an acceptable claim.
\end{example}

So by having appropriate conditions in the antecedent of a simple rule we can disable the rule by generating a counterargument that attacks it. This in effect stops the usage of the simple rule. This means that we have a convention to attack an argument based on the inferences obtained by the simple logic (e.g. as in 
Example \ref{ex:counterarguments:simple:metro1} and Example \ref{ex:counterarguments:simple:gov1}), or on the rules used (e.g. 
Example \ref{ex:counterarguments:simple:metro2}). 

This way to disable rules by adding appropriate conditions (as in Example \ref{ex:counterarguments:simple:metro2}) is analogous to the use of abnormality predicates in formalisms such as circumscription (see for example \cite{Macarthy80}). We can use the same approach to capture defeasible reasoning in other logics such as classical logic (as for example, the use of the ${\tt ok}$ predicate in the arguments in Figure \ref{fig:classical:medical}). Note, this does not mean that we turn the base logic into a nonmonotonic logic. Both simple logic and classical logic are monotonic logics. 
Hence, for a simple logic knowledgebase $\Delta$ (and similarly for a classical logic knowledgebase $\Delta$), the set of simple arguments (respectively classical arguments) obtained from $\Delta$ is a subset of the set of simple arguments (respectively classical arguments)  obtained from $\Delta \cup \{\alpha\}$ where $\alpha$ is a formula not in $\Delta$. 
But at the level of evaluating arguments and counterarguments, we have non-monotonic defeasible behaviour as illustrated by Example \ref{ex:counterarguments:simple:metro2}. 

In the section, we have focused on using simple logic as a base logic. But it is very weak since it only has modus ponens as a proof rule. 
There is a range of logics between simple logic and classical logic called conditional logics. 
They are monotonic, and they capture interesting aspects of defeasible reasoning.
We will briefly consider one type of conditional logic in Section \ref{section:conditional}.

\subsection{Other approaches to structured argumentation}

Other approaches to structured argumentation such as ASPIC+ \cite{Prakken10,MP14} and assumption-based argumentation (ABA) \cite{Toni2013,Toni14} also have an argument construction process that is monotonic but an argument evaluation process that is non-monotonic.
These approaches apply rules to the formulae from the knowledge base, where the rules may be defeasible. These rules are defeasible in the sense they describe defeasible knowledge but the underlying logic is essentially modus ponens (i.e. analogous to simple logic as investigated for deductive argumentation) and so monotonic. Note, in these rule-based approaches, an argument is seen as a tree whose root is the claim or conclusion, whose leaves are the premises on which the argument is based, and whose structure corresponds to the application of the rules from the premises to the conclusion.

Through the dialectical semantics of argumentation, it is possible to formalize non-monotonic logics. Bondarenko {\em et. al.} showed how assumption-based argumentation subsumes a range of key non-monotonic logics including Theorist, default logic, logic programming, autoepistemic logic, non-monotonic modal logics, and certain instances of circumscription as special cases \cite{Bon97}. 
The use of assumptions can be introduced to other approaches to structured argumentation (see for example, for ASPIC+ \cite{Prakken10}).

In defeasible logic programming (DeLP) \cite{GS04,GS14}, another approach to structured argumentation, strict and defeasible rules are used in a form of logic programming. The language includes a default negation (i.e. a form of negation-as-failure) which gives a form of non-monotonic behaviour. Essentially, negation-as-failure means that if an atom cannot be proved to be true, then assume it is false. For example, the following is a defeasible rule in DeLP where the negated atom $\sim\hspace{-2mm}{\tt ~crossRailwayTracks}$ is the consequent, $-\hspace{-2mm} <$ is the defeasible implication symbol, and $not \sim\hspace{-1mm}{\tt trainIsComing}$ is the condition. Furthermore, $\sim\hspace{-1mm}{\tt trainIsComing}$ is a negated atom, and $not$ is the default negation operator. 
\[
\sim\hspace{-2mm}{\tt ~crossRailwayTracks} \; -\hspace{-2mm} < \; not \sim\hspace{-1mm}{\tt trainIsComing}
\] 
For the above defeasible rule, if we cannot prove $\sim\hspace{-1mm}{\tt trainIsComing}$ (i.e. we cannot show that the train is not coming), then  $not \sim\hspace{-1mm}{\tt trainIsComing}$ holds, and therefore we infer $\sim\hspace{-2mm}{\tt ~crossRailwayTracks}$ (i.e. we shouldn't cross the tracks). 

In DeLP, the negation-as-failure is with respect to all the strict knowledge (i.e. the subset of the knowledge that is assumed to be true), rather than just the premises of the argument, and this means that as formulae are added to the knowledgebase, it may be necessary to withdraw arguments. For instance, if there is no strict knowledge, then we can construct an argument that has the above defeasible rule as the premise, and $\sim\hspace{-2mm}{\tt ~crossRailwayTracks}$ as the claim. But, if we then add $\sim\hspace{-1mm}{\tt trainIsComing}$ to the knowledgebase as strict knowledge, then we have to withdraw this argument. So unlike deductive argumentation, and the other approaches to structured argumentation, DeLP is not monotonic in argument contruction. However, like deductive argumentation, and the other approaches to structured argumentation, DeLP is non-monotonic in argument evaluation.

\section{Default logic}
\label{section:default}

As a basis of representing and reasoning with default knowledge,
default logic, proposed by Reiter \cite{Reiter1980}, is one of the best known and most widely studied
formalisations of default reasoning. Furthermore, it offers a very expressive and
lucid language.  In default logic, knowledge is represented as a {\em
default theory}, which consists of a set of first-order formulae and a
set of {\em default rules} for representing default
information. Default rules are of the following form, where $\alpha$,
$\beta$ and $\gamma$ are classical formulae.

\[
{\alpha : \beta \over \gamma}
\]

The inference rules are those of classical logic plus a special
mechanism to deal with default rules: Basically, if $\alpha$ is
inferred, and $\neg\beta$ cannot be inferred, then infer $\gamma$. For
this, $\alpha$ is called the pre-condition, $\beta$ is called the
justification, and $\gamma$ is called the consequent.

\subsection{Inferencing in default logic}

Default logic is an extension of classical logic. Hence, all classical inferences from the classical information in a default theory are derivable (if there is an extension). The default theory then augments these classical inferences by default inferences derivable using the default rules.

\begin{definition}
\label{defi:extension}
Let $(D,W)$ be a default
theory, where D is a set of default rules and W is a set of classical
formulae. Let $Cn$ be the function that for a set of formulae
returns the set of classical consequences of those formulae. 
The operator $\Gamma$ indicates what
conclusions are to be associated with a given set $E$ of formulae,
where $E$ is some set of classical formulae. 
For this,
$\Gamma(E)$ is the smallest set of classical formulae such that the
following three conditions are satisfied. 

\begin{enumerate}

\item $W \subseteq \Gamma(E)$

\item $\Gamma(E) = Cn(\Gamma(E))$

\item For each default in $D$, where $\alpha$ is the pre-condition, $\beta$ is the justification, and $\gamma$ is the consequent, the following holds: 

\[
\mbox{if } \alpha \in \Gamma(E), 
\mbox{and }  \neg \beta \not \in E,   
\mbox{then } \gamma \in \Gamma(E)
\]

\end{enumerate}

We refer to $E$ as the satisfaction set, and $\Gamma(E)$ the putative extension.
\end{definition}

Once $\Gamma(E)$ has been identified, $E$ is an extension of $(D,W)$ iff $E=\Gamma(E)$. If $E$ is an extension, then the first condition ensures that the set of classical formulae $W$ is also in the extension, the second condition ensures the
extension is closed under classical consequence, and the third
condition ensures that for each default rule, if the pre-condition is
in the extension, and the justification is consistent with the
extension, then the consequent is in the extension. 

We can view $E$ as the set of formulae for which we are ensuring consistency with the justification of each default rule that we are attempting to apply. We can view $\Gamma(E)$ as the set of putative conclusions of a default theory: It contains $W$, it is closed under classical consequence, and for each default that is applicable (i.e. the precondition is $\Gamma(E)$ and the justification is satisfiable with $E$), then the consequent is in $\Gamma(E)$. We ask for the smallest $\Gamma(E)$ to ensure that each default rule that is applied is grounded. This means that it is not the case that one or more default rules are self-supporting. For example, a single default rule is self-supporting if the pre-condition is satisfied using the consequent.  
The test $E$ = $\Gamma(E)$ ensures that the set of formulae for which the justifications are checked for consistency coincides with the set of putative conclusions of the default theory.  If $E \subset \Gamma(E)$, then not all applied rules had their justification checked with $\Gamma(E)$. If $\Gamma(E) \subset E$, then the rules are checked with more than is necessary.

\begin{example}
\label{ex:tweety}
Let $D$ be the following set of defaults.
\[
\frac{\tt bird(X) : \neg penguin(X) \wedge fly(X)}{\tt fly(X)}
\]
\[
\frac{\tt penguin(X) : bird(X)}{\tt bird(X)}
\hspace{1cm}
\frac{\tt penguin(X) : \neg fly(X)}{\tt \neg fly(X)}
\]
For $(D,W)$, where  $W$ is $\{\tt bird(Tweety) \}$, we obtain one extension 
\[
Cn(\{\tt bird(Tweety), fly(Tweety) \})
\]
For $(D,W)$, where $W$ is $\{\tt penguin(Tweety) \}$, we obtain one extension 
\[
Cn(\{\tt penguin(Tweety), bird(Tweety), \neg fly(Tweety)\})
\]
\end{example}

\begin{figure}[t]
\begin{center}
\begin{tabular}{|c|c|c|p{4cm}|}
\hline
&&&\\
Attempt & $E$ & $\Gamma(E)$ & Extension? \\
&&&\\
\hline
\hline
1 
& $Cn(\{{\tt bird(Tweety)}\})$ 
& {\tt bird(Tweety)} 
& $E\subset\Gamma(E)$\\
&& {\tt fly(Tweety)}&\\
\hline
2
& $Cn(\{{\tt fly(Tweety)} \})$
& {\tt bird(Tweety)} 
& $E\subset\Gamma(E)$\\
&&{\tt fly(Tweety)} &\\
\hline
3
& $Cn(\{{\tt bird(Tweety)}, $
& {\tt bird(Tweety)} 
& $E = \Gamma(E)$\\
& ${\tt fly(Tweety)} \})$&{\tt fly(Tweety)}&\\
\hline
4
& $Cn(\{\tt bird(Tweety),$
& {\tt bird(Tweety)} 
& $\Gamma(E)\subset E$\\
&$\neg\mbox{\tt fly(Tweety)} \})$ &&\\
\hline
5
& $Cn(\{\neg\mbox{\tt bird(Tweety)},$ 
& ${\tt bird(Tweety)} \}$
& $\Gamma(E)\not\subseteq E$ \& $E\not\subseteq\Gamma(E)$\\
&$\neg\mbox{\tt fly(Tweety)} \})$ &&\\
\hline
\end{tabular}
\end{center}
\caption{A non-exhaustive number of attempts are made for determining an extension. In each attempt, a guess is made for $E$, and then $\Gamma(E)$ is calculated.}
\end{figure}

Possible sets of conclusions from a default theory are given in terms
of extensions of that theory. A default theory can possess
multiple extensions because different ways of resolving conflicts
among default rules lead to different alternative extensions. For
query-answering this implies two options: in the credulous
approach, we accept a query if it belongs to one of the extensions of
a considered default theory, whereas in the skeptical
approach, we accept a query if it belongs to all extensions of the
default theory. 

The notion of extension in default logic overlaps with the notion of extension in argumentation. However, there is a key difference between the two notions. In the former, the extensions are derived directly from reasoning with the knowledge, whereas in the latter, the extensions are derived from the instantiated argument graph, and that graph has been generated from the knowledge. As a consequence, default logic suppresses the inconsistency arising within the relevant knowledge whereas argumentation draws the inconsistency out in the form of arguments and counterarguments.

\subsection{Using default logic in deductive argumentation}

We can use default logic as a base logic. Normally, a base logic is monotonic. However, there is no reason to not use a non-monotonic logic such as default logic.
Let $\vdash_d$ be the consequence relation for default logic. For a default theory $\Delta = (D,W)$, and a propositional formula $\alpha$, $\Delta\vdash_d\alpha$ denotes that $\alpha$ is in a default logic extension of $(D,W)$. So the consequence relation is credulous. We could use a skeptical version as an alternative.

\begin{definition}
For a  default theory $\Phi = (D,W)$,
and a classical formula $\alpha$,
$\langle \Phi, \alpha \rangle$ is a {\bf default argument}
iff $\Phi \vdash_d \alpha$ 
and there is no proper subset $\Phi'$ of $\Phi$ 
such that $\Phi' \vdash_d \alpha$. 
\end{definition}

For the above definition, $\Phi' = (D',W')$ is a proper subset of $\Phi = (D,W)$
iff $D' \subseteq D$ and $W' \subset W$
or $D' \subset D$ and $W' \subseteq W$.

We now consider a definition for attack 
where the counterargument attacks an argument when it negates the justification of a default rule in the premises of the argument. 
For this we require a function ${\sf DefaultRules}(A)$ to return the default rules used in the premises of an default argument.

\begin{definition}
For arguments $A$ and $B$, where $A$ is a default argument, and $B$ is either a default argument or a classical argument, $B$ {\bf justification undercuts} $A$ 
if there is $\alpha:\beta/\gamma \in {\sf DefaultRules}(A)$ 
s.t. ${\sf Claim}(B) \vdash \neg \beta$.
\end{definition}

\begin{figure}[t]
\begin{center}
\begin{tikzpicture}[->,>=latex,thick]
\node (a1) [text centered,text width=6cm,shape=rectangle,fill=yellow,draw] at (4,4) {
\[
\begin{array}{l}
\mbox{\tt bird(Tweety) }\\
\left(\frac{\mbox{\tt bird(X)}:
\neg \mbox{\tt penguin(X)} \wedge \mbox{\tt fly(X)}}
{ \mbox{\tt fly(X)}} \right)\\
\hline
\mbox{\tt fly(Tweety)} \\
\end{array}
\]
};
\node (a2) [text centered,text width=3cm,shape=rectangle,fill=yellow,draw] at (0,1) {
\[
\begin{array}{l}
\mbox{\tt penguin(Tweety) }\\
\hline
\mbox{\tt penguin(Tweety) }
\end{array}
\]
};
\node (a3) [text centered,text width=6cm,shape=rectangle,fill=yellow,draw] at (6,1) {
\[
\begin{array}{l}
\mbox{\tt penguin(Tweety) }\\
\left(\frac{\mbox{\tt penguin(X)}:
\neg \mbox{\tt fly(X)}}
{ \neg \mbox{\tt fly(X)}} \right)\\
\hline
\neg \mbox{\tt fly(Tweety) }
\end{array}
\]
};
\path	(a2) edge node[] {} (a1);
\path	(a3) edge node[] {} (a1);
\end{tikzpicture}
\end{center}
\caption{\label{fig:defaultlogic}Example of using default logic as a base logic in deductive argumentation. Here the top argument and the right arguments are default arguments, whereas the left argument is a classical argument. Each argument is presented with the premises above the line, and the claim below the line. Each default rule is given in brackets.}
\end{figure}
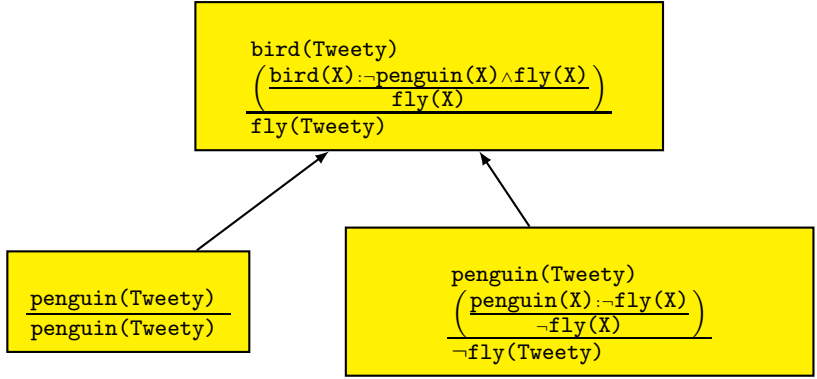

We illustrate justification undercuts in Figure \ref{fig:defaultlogic}. In the figure, the left undercut involves a classical argument, and the right undercut involves a default argument.

Using default logic as a base logic in this way does not affect the argument contruction being monotonic. Adding a formula to the knowledgebase would not cause an argument to be withdrawn. Rather it may allow further arguments to be constructed.

The advantage of using default logic in arguments is that it allows default inferences to be drawn. 
This means that we use a well-developed and well-understand formalism for representing and reasoning with the complexities of default knowledge. Hence, we can have a richer and more natural representation of defaults. 
It also means that inferences can be drawn in the absence of reasons to not draw them. For instance, in Figure \ref{fig:defaultlogic}, we can conclude ${\tt fly(Tweety)}$ from ${\tt bird(Tweety)}$ in the absence of knowing whether it is a penguin. In other words, we just need to know that it is consistent to believe that it is not a penguin and that it is consistent to believe that it can fly. Furthermore, we just need to do this consistency check within the premises of the argument. Note, this consistency check is different to the consistency check used for default negation in DeLP which involves checking consistency with all the strict knowledge (i.e. the subset of knowledge that is assumed to be correct) \cite{GS04}. 

In this paper, we have only given an indication of how default logic can be used to capture aspects of non-monotonic reasoning (for a comprehensive review of default logic, see \cite{Besnard1989}) in deductive argumentation. There are various ways (e.g. by Santos and Pa\~{v}ao Martins \cite{Santos2008}) that default logic could be harnessed in deductive argumentation to give richer behaviour . Also possible definitions for counterarguments could be adapted from an approach to argumentation based on default logic by Prakken \cite{Prakken1993}.

\section{KLM logics for non-monotonic reasoning}
\label{section:conditional}

The KLM logics for non-monotonic reasoning are a family of conditional logics developed by Kraus, Lehmann and Magidor  \cite{KLM90} to capture aspects of non-monotonic reasoning and where each logic has a proof theory and a possible worlds semantics. 
We focus here on a member of this family called System P which has the language composed of formulae of the form $\alpha\Rightarrow\beta$ where $\alpha,\beta$ are propositional formulae, and it has the the following set of proof rules. 

\[
(REF) \; \frac{}{A \Rightarrow A} 
\hspace{1cm} (CUT) \; \frac{A \Rightarrow B \hspace{4mm} A \wedge B \Rightarrow C}{A \Rightarrow C}
\]

\[
(LLE) \; \frac{A \equiv B \hspace{4mm} A \Rightarrow C}{B \Rightarrow C}
\hspace{1cm} (RW) \; \frac{ \vdash A \rightarrow B \hspace{4mm} C \Rightarrow A}{C \Rightarrow B}
\]

\[
(AND) \frac{A \Rightarrow B \hspace{4mm}  A \Rightarrow C }{ A \Rightarrow B \wedge C}
\hspace{5mm} (OR) \; \frac{A \Rightarrow C \hspace{4mm}  B \Rightarrow C}{A \vee B \Rightarrow C}
\]

\[
(CM) \; \frac{A \Rightarrow B \hspace{4mm}  A \Rightarrow C}{A \wedge B \Rightarrow C} 
\]

\[
(LOOP) \; \frac{A_0 \Rightarrow A_1 \hspace{4mm} A_1 \Rightarrow A_2
\ldots A_{k-1} \Rightarrow A_k \hspace{4mm} A_k \Rightarrow A_0 }{ A_0 \Rightarrow A_k }
\]

We illustrate this proof systems using the following examples of inferences. 
In the examples, we can see how the reasoning is monotonic in that no inferences (i.e. no formula of the form $\alpha\Rightarrow\beta$) are retracted.
However, within these formulae, the defeasible reasoning is encoded. For instance, in Example \ref{ex:klm1}, we have a formula that says ``birds fly", and we have the inference that says ``birds that are penguins do not fly".

\begin{example}
\label{ex:klm1}
From the following statements
\[
\begin{array}{lll}
\mbox{\tt penguin} \Rightarrow \mbox{\tt bird}
\hspace{1cm}
& \mbox{\tt penguin} \Rightarrow \neg \mbox{\tt fly}
\hspace{1cm}
& \mbox{\tt bird} \Rightarrow \mbox{\tt fly}
\end{array}
\]
we get the following inferences
\[
\begin{array}{ll}
\mbox{\tt penguin} \wedge \mbox{\tt bird} \Rightarrow \neg \mbox{\tt fly}
\hspace{1cm}
& \mbox{\tt fly} \Rightarrow \neg \mbox{\tt penguin}\\
\mbox{\tt bird} \Rightarrow \neg \mbox{\tt penguin}
\hspace{1cm}
& \mbox{\tt bird} \vee \mbox{\tt penguin} \Rightarrow \mbox{\tt fly}\\
\mbox{\tt bird} \vee \mbox{\tt penguin} \Rightarrow \neg \mbox{\tt penguin}
\hspace{1cm}
&\\
\end{array}
\]
\end{example}



We can use System P as a base logic in a deductive argumentation system.
We start by giving the following definition of an argument.

\begin{definition}
Conditional logic argument
For a set of conditional statements $\Phi$
and a set of propositional formulae $\Psi$,
if $\Phi\vdash_P\alpha\Rightarrow\beta$ and $\wedge\Psi\equiv\alpha$,
then a {\bf preferential argument} is
$\langle \Phi\cup\Psi, \alpha\Rightarrow\beta \rangle$.
\end{definition}

\begin{example}
For premises $\Phi = \{ \mbox{\tt penguin} \Rightarrow \mbox{\tt bird}, \mbox{\tt penguin} \Rightarrow \neg \mbox{\tt fly} \}$
and $\Psi = \{ \mbox{\tt penguin}, \mbox{\tt bird} \}$,
the following is a preferential argument.
\[
\langle \Phi\cup\Psi, \mbox{\tt penguin} \wedge  \mbox{\tt bird} \Rightarrow \neg \mbox{\tt fly} \rangle
\]
\end{example}

Next, we can define a range of attack relations. 
The following definition is not exhaustive as there are further options that we could consider.

\begin{definition}
Some preferential attack relations:
Let $A_1 = \langle \Phi\cup\Psi, \alpha\Rightarrow\beta \rangle$ 
and $A_2 = \langle \Phi'\cup\Psi', \gamma\Rightarrow\delta \rangle$. 
\begin{itemize}
\item $A_2$ is a {\bf rebuttal} of $A_1$ iff $\delta \vdash \neg \beta$ and $\gamma\vdash\alpha$
\item $A_2$ is a {\bf direct rebuttal} of $A_1$ iff $\delta \equiv \neg \beta$ and $\gamma\vdash\alpha$
\item $A_2$ is a {\bf undercut} of $A_1$ iff $\delta \vdash \neg \alpha$
\item $A_2$ is a {\bf canonical undercut} of $A_1$ iff $\delta \equiv \neg \alpha$
\item $A_2$ is a {\bf direct undercut} of $A_1$ iff there is $\sigma \in \Psi$ such that $\delta \equiv \neg \sigma$
\end{itemize}
\end{definition}

\begin{example}
For $A_1$ and $A_2$ below, $A_2$ is a direct rebuttal of $A_1$, but not vice versa,
\begin{itemize}
\item $A_1 = \langle \Phi_1\cup\Psi_1, \mbox{\tt bird} \Rightarrow \mbox{\tt fly} \rangle$
\item $A_2 = \langle \Phi_2\cup\Psi_2, \mbox{\tt penguin} \wedge \mbox{\tt bird} \Rightarrow \neg \mbox{\tt fly} \rangle$
\end{itemize}
where 
\begin{itemize}
\item $\Phi_1 = \{ \mbox{\tt bird} \Rightarrow \mbox{\tt fly} \}$,
\item $\Psi_1 = \{ \mbox{\tt bird} \}$,
\item $\Phi_2 = \{ \mbox{\tt penguin} \Rightarrow \mbox{\tt bird}, 
	\mbox{\tt penguin} \Rightarrow \neg \mbox{\tt fly} \}$,
\item $\Psi_2 = \{ \mbox{\tt penguin}, \mbox{\tt bird} \}$.
\end{itemize}
\end{example}

\begin{example}
For $A_1$ and $A_2$ below, $A_2$ is a direct rebuttal of $A_1$, but not vice versa,
\begin{itemize}
\item $A_1 = \langle \Phi_1\cup\Psi_1, \mbox{\tt matchIsStruck} \Rightarrow \mbox{\tt matchLights} \rangle$
\item $A_2 = \langle \Phi_2\cup\Psi_2, \mbox{\tt matchIsStruck} \wedge \mbox{\tt matchIsWet} \Rightarrow \neg \mbox{\tt matchLights} \rangle$
\end{itemize}
\end{example}

Harnessing System P, and the other members of the KLM family, offers the ability to undertake intuitive reasoning with defeasible rules. This allows for plausible consequences from knowledge to be investigated. It also allows for more efficient representation of knowledge to be undertaken since fewer rules would be required when compared with using simple logic. Furthermore, this reasoning can be implemented using automated reasoning \cite{Giordano2009}.

\section{Further conditional logics}

Conditional logics are a valuable alternative to classical logic for knowledge representation and reasoning. Whilst many conditional logics extend classical logic, the implication introduced is normally more restricted than the strict implication used in classical logic. This means that many knowledge modelling situations, such as for non-monotonic reasoning, can be better captured by conditional logics (such as \cite{Delgrande87,KLM90,Arlo92}). 

By using conditional logic as a base logic, we have a range of options for more effective modelling complex real-world scenarios. 
For instance, they can be used to capture hypothetical statements of the form ``If $\alpha$ were true, then $\beta$ would be true". This done by introducing an extra connective $\Rightarrow$ to extend a classical logic language. Informally, $\alpha\Rightarrow\beta$ is valid when $\beta$ is true in the possible worlds where $\alpha$ is true.  Representing and reasoning with such knowledge in argumentation is valuable because useful arguments exist that refer to fictitious and hypothetical situations as shown by Besnard {\em et. al.} \cite{BGR13}.

So we are proposing that we should move beyond simple logic (i.e. modus ponens) that is essentially the logic used for most structured argumentation systems. 
Even though formalisms such as ASPIC+ and ABA are general frameworks that accept a wide range of proof systems, most exmaples of using the frameworks involve simple rule-based systems (i.e. rules with modus ponens).

Once we accept that we can move beyond simple rule-based systems, we have a wide range of formalisms that we could consider, in particular conditional logics. This then raises the question of what proof theory do we want for reasoning with the defeasible rules. Central to such considerations is whether we want to have contrapositive reasoning. 

In considering this issue for argumentation, Caminada recalls two simple examples where the inference of contrapositives is problematic \cite{Caminada2008}.

\begin{itemize}
\item Men usually do not have beards. But, this does not imply that if someone does have a beard, then that person is not a man.
\item If I buy a lottery ticket, then I will normally not win a prize. But, this does not imply that if I do win a prize, then I did not buy a ticket.
\end{itemize}

To identify situation where contrapositive reasoning is potentially desirable, Caminada described the following two types of situation. 

\begin{description}
\item[Epistemic] Defeasible rules describe how certain facts hold in relation to each other in the world. So the world exists independently of the rules. Here, contrapositive reasoning may be appropriate.
\item[Constitutive] Defeasible rules (in part) describe how the world is constructed (e.g. regulations). So the world does not exist independently of the rules.  Here contrapositive reasoning is not appropriate.
\end{description}

The first example below illustrates the first kind of situation, and the second example illustrates the second kind of situation. Note that the knowledge in each example is syntactically identical. 

\begin{example}
The following rules describe an epistemic situation.
From these rules, it would be reasonable to infer $\tt \neg A$ from $\tt L$.
\begin{itemize}
\item $\tt O \Rightarrow A$ - goods ordered three months ago will probably have arrived now.
\item $\tt A \Rightarrow C$ - arrived goods will probably have a customs document.
\item $\tt L \Rightarrow \neg C$ - goods listed as unfulfilled will probably lack customs document.
\end{itemize}
\end{example}

\begin{example}
The following rules describe a constitutive situation.
From these rules, it would not be reasonable to infer $\tt \neg M$ from $\tt P$.
\begin{itemize}
\item $\tt S \Rightarrow M$ - snoring in the library is form of misbehaviour.
\item $\tt M \Rightarrow R$ - misbehaviour in the library can result in removal from the library.
\item $\tt P \Rightarrow \neg R$ - professors cannot be removed from the library.
\end{itemize}
\end{example}




A wide variety of conditional logics have been proposed to capture various aspects of conditionality including deontics, counterfactuals, relevance, and probability \cite{Chellas1975}. These give a range of proof theories and semantics for capturing different aspects of reasoning with conditions, and many of these do not support contrapositive reasoning.

\section{Modelling defeasible knowledge}
\label{section:modelling}

Another important aspect of non-monotonic reasoning is how we model defeasible knowledge in a meaningful way.
A defeasible or default rule is a rule that is generally true but has exceptions and so is sometimes untrue.
But this explanation still leaves some latitude as to what we really mean by a defeasible rule. To illustrate this issue, consider the following rule.

\[
{\tt birds \; fly \hspace{5mm} (*)}
\]
If we take all the {\bf normal} situations (or observations, or worlds or days), and in the majority of these birds fly, then (*) may be equal to the following
\[
{\mbox{\tt birds normally fly}}
\]
Or if we take the set of all birds  (or all the birds you have seen, or read about, or watched on TV) and the {\bf majority} of this set fly, then (*) may be equal to the following
\[
{\mbox{\tt most birds fly}}
\]
Or perhaps if we take the set of all birds, we know the majority have the {\bf capability} to fly, then (*) could equate with the following. --- 
though this in turn raises the question of what it is to know something has a capability
\[
{\mbox{\tt most birds have the capability to fly}}
\]
Or if we take an {\bf idealized} notion of a bird that we in a society are happy to agree upon, then we could equate (*) with the following
\[
{\mbox{\tt a prototypical bird flies}}
\]

As another illustration of this issue, consider the following rule which is in the vein as the ``birds fly" example but introduces further complications. 

\[
{\mbox{\tt birds lay eggs} \hspace{5mm} (**)}
\]
We may take (**) as meaning the following --- but on a given day, (or given situation, or observation) a given bird bird will probably not lay an egg.
\[
{\mbox{\tt birds normally lay eggs}}
\]
Or we could say (**) means the following --- but half the bird population is male and therefore don't lay eggs. 
\[
{\mbox{\tt most birds have the capability lay eggs}}
\]
Or perhaps we could say (**) means the following --- where 
\[
{\mbox{\tt most species of bird reproduce by laying eggs}}
\]
However, the above involves a lot of missing information to be filled in, and in general it is challenging to know how to formalize a defeasible rule.

Obviously different applications call for different ways to interpret defeasible rules. Specifying the underlying ontology by recourse to description logic may be useful, and in some situations formalization in a probabilistic logic (see for example, \cite{Bacchus1990}) may be appropriate 

\section{Conclusions}
\label{section:conclusions}

Discrimination between the monotonic and non-monotonic aspects of a deductive argumentation system is important to better understand the nature of deductive argumentation. Each  deductive argumentation system is based on a base logic which can be monotonic or non-monotonic. In either case, the construction of arguments and counterarguments is monotonic in the sense that adding knowledge to the knowledgebase may increase the set of arguments and counterargument but it cannot reduce the set of arguments or counterarguments. However, at the evaluation level, deductive argumentation is non-monotonic since adding arguments and counterarguments to the instantiated graph  may cause arguments to be withdrawn from an extension, or even for extensions to be withdrawn.

Defeasible formulae (such as defeasible rules) are an important kind of knowledge in argumentation, as in non-monotonic reasoning. They are formulae that are often correct, but sometimes can be incorrect. Depending on what aspect of defeasible knowledge we want to model, there is a wide variety of base logics that we can use to represent and reason with it. This range includes simple logic and classical logic when we use appropriate conventions such abnormality predicates, in which case, we can overturn an argument based defeasible knowledge by recourse counterarguments that attack the assumption of normality. This range of base logics also includes default logic and conditional logics.


\newcommand{\etalchar}[1]{$^{#1}$}

\end{document}